\documentclass[journal]{IEEEtran}

\usepackage{amsmath,amssymb,amsfonts}
\usepackage{mathtools}
\usepackage{graphicx}
\usepackage{booktabs}
\usepackage{multirow}
\usepackage{tikz}
\usepackage{float}
\usepackage{bm}
\usepackage{xcolor}        
\usepackage[hidelinks]{hyperref}
\usepackage{url}
\DeclareMathAlphabet{\mathbbold}{U}{bbold}{m}{n}
\usepackage[caption=false,font=footnotesize]{subfig}
\usepackage{placeins}

\newcommand{\citet}[1]{\cite{#1}}
\newcommand{\citep}[1]{\cite{#1}}

\title{Optimized Instance Alteration for Explaining and Assessing Robustness of Classifiers}

\author{Evgenii~Kuriabov,
        David~Miller,
        and~Jia~Li,~\IEEEmembership{Fellow,~IEEE}
\thanks{E. Kuriabov and J. Li are with the Department of Statistics, The Pennsylvania State University, University Park, PA 16802, USA. Emails: \texttt{eak5582@psu.edu}, \texttt{jiali@psu.edu} }.
\thanks{D. Miller is with the Department of Electrical Engineering, The Pennsylvania State University, University Park, PA 16802, USA. Email: \texttt{djm25@psu.edu}}
\thanks{The research of E. Kuriabov and J. Li has been supported by the NSF under Award No. 2205004.}
\thanks{Correspondence: Evgenii Kuriabov.}
}

\begin{document}
\maketitle

\begin{abstract}
In this work, we propose a unified approach for diagnosing misclassification and assessing the robustness of black-box classifiers. Central to our method is an optimization framework that modifies an instance so that the classifier predicts a specified target label, while ensuring that the modification remains easily explainable. The objective function contains two components: an explainability-aware $L_0$ (XA-$L_0$) penalty that promotes sparse and interpretable modifications, and a classifier loss objective that steers the perturbed instance toward the desired output. This integrated optimization formulation is used both to identify the underlying causes of misclassification and to evaluate robustness by determining how an instance can change within a tolerance region before being reassigned to another class.
To quantify robustness, we introduce the Tolerance Region Confusion Matrix (TOR-Confusion Matrix), which measures a classifier’s susceptibility by modeling the class-to-class transition probabilities induced by tolerance-bounded perturbations. We validate the proposed method on both image and tabular datasets, demonstrating its ability to jointly deliver interpretability and robustness assessment.
\end{abstract}


\section{Introduction}

Interpretable machine learning has become essential for understanding and validating model decisions in domains such as healthcare, finance, and autonomous systems. 
Despite impressive predictive performance, modern deep models often act as opaque black boxes, offering little insight into why a specific prediction was made or how a misclassification could be corrected. 
This opacity limits trust, accountability, and deployment in high-stakes environments where the reasoning behind predictions must be transparent. 
A principled interpretability framework should therefore not only explain a model’s decision but also suggest how the decision could be altered through minimal, meaningful changes to the input. For example, in credit or insurance approval, a counterfactual explanation can indicate that only a modest increase in credit score or reported income would flip a rejection to an approval, providing actionable feedback to an applicant. As another example, in scientific or clinical settings where features are measured with finite precision, an explanation may reveal that the predicted label changes under a perturbation smaller than the typical measurement resolution of a key feature, suggesting that this feature should be measured more precisely in practice or that the decision is unstable to measurement noise.

Counterfactual explanations offer a natural and intuitive mechanism for achieving this goal. 
Given a misclassified instance, the objective is to identify the smallest modification that would change the model’s output to the desired label. Likewise, given a correctly classified instance, the objective is to identify how close the instance is to being misclassified.
Existing approaches typically encourage sparsity using $L_0$-style penalties or their relaxations~\cite{wachter2017counterfactual,dhurandhar2018desiderata,dandl2020multiobjective}, seeking to minimize the number of altered features. However, sparsity alone does not guarantee interpretability.
An $L_0$ penalty constrains the number of modified features, but not the nature or coherence of those changes.
In image domains, enforcing sparsity may result in isolated pixels whose alteration creates implausible visual artifacts.
In tabular settings, sparsity that ignores inherent feature relationships produces edits that are numerically small but conceptually nonsensical.
Thus, while $L_0$ promotes minimality, it falls short of ensuring that the resulting counterfactuals are actually interpretable, a gap we directly address in our formulation.
Sampling-based inverse classification methods, such as Growing Spheres~\cite{laugel2017inverse}, further struggle with scalability and often neglect dependencies among variables, resulting in disjoint or unstable counterfactuals.

Counterfactual generation is also closely related to methods for constructing adversarial examples and test-time evasion attacks, which similarly search for input modifications that induce a change in a model’s prediction~\cite{goodfellow2015explaining,madry2018towards,cohen2019certified}. The key distinction lies in intent and structure: adversarial methods typically aim to produce imperceptible or noise-like perturbations that change the decision while remaining difficult to detect, whereas counterfactual explanations seek coherent, semantically meaningful modifications that reveal how and why a decision could change. This contrast motivates our emphasis on structured, interpretable perturbations—changes that are aligned with domain structure rather than optimized solely for minimal norm or visual imperceptibility.

To address these limitations, we propose a unified optimization framework that generates \emph{interpretable corrections} by jointly enforcing correctness, parsimony, and semantic coherence.
The approach integrates three complementary components: a classification term that enforces the desired label, a proximity regularizer that maintains realism, and a novel \emph{Explainability-Aware $L_0$ (XA-$L_0$)} penalty that promotes structured sparsity by coupling related features.
For tabular data, this coupling captures meaningful feature dependencies, such as correlations or learned communities in a feature graph, while for image data, an edge-aware spatial regularizer favors clustered modifications near object boundaries, where changes are most visually salient, avoiding the scattered or visually implausible pixel changes produced by standard sparsity penalties.
Together, these components yield compact and semantically consistent corrections that respect both statistical relationships and perceptual structure, providing interpretable pathways for refining model predictions.

Beyond generating individual corrections, we introduce the \emph{Tolerance-Region Confusion Matrix (TOR-CM)} to quantify model robustness under interpretable perturbations. 
Unlike conventional adversarial or certified robustness measures~\cite{goodfellow2015explaining,madry2018towards,cohen2019certified}, which evaluate sensitivity to infinitesimal or norm-bounded noise, TOR-CM characterizes how predictions change within bounded, semantically meaningful tolerance regions defined by the XA-$L_0$ geometry. 
This perspective links interpretability and robustness within a single optimization-based framework, enabling a unified analysis of both model behavior and its stability under human-understandable modifications.

The proposed framework thus unifies counterfactual interpretability and robustness assessment through a structured optimization objective. 
Extensive experiments on image and tabular datasets show that XA-$L_0$ generates coherent, domain-aligned corrections while providing faithful, interpretable explanations of the classifier’s decisions.
Moreover, the TOR-CM analysis reveals that models differ not only in accuracy but also in their intrinsic stability under interpretable perturbations, providing a more complete characterization of model reliability in real-world settings.

\section{Preliminaries and Related Work}
\label{sec:related}

Counterfactual explanations have emerged as one of the most intuitive forms of post-hoc interpretability, aiming to explain or correct a model’s decision by identifying minimal input changes that would alter its predicted class. Given an instance $\mathbf{x}^{(o)} \in \mathbb{R}^d$ and a trained classifier $f(\mathbf{x})$, the objective is to find a perturbed point $\mathbf{x}^*$ such that $f(\mathbf{x}^*) \neq f(\mathbf{x}^{(o)})$ while keeping $\mathbf{x}^*$ as similar and plausible as possible. This formulation offers a human-aligned explanation by revealing ``what needs to change'' to achieve a desired outcome and, correspondingly, ``which features contribute to the misclassification.'' Likewise, it can identify how close a correctly classified instance is to being misclassified, and which feature changes would result in such misclassification.

Recent surveys~\cite{verma2024counterfactual,karimi2021algorithmic,10.1145/3677119} highlight that counterfactual frameworks differ not only in their technical formulation but also in their underlying goals. Some prioritize \textit{classification fidelity}, ensuring that the modified instance reliably flips the prediction~\cite{wachter2017counterfactual,dhurandhar2018desiderata}; others emphasize \textit{proximity and feasibility}, constraining changes to remain close to the original input. Sparsity-based methods focus on \textit{parsimony}, seeking to alter as few features as possible~\cite{dandl2020multiobjective,mothilal2020explaining}, while generative and manifold-based approaches~\cite{antoran2020getting,joshi2019towards,van2021interpretable} aim for \textit{plausibility}, producing counterfactuals that remain within the natural data distribution. More recent causal formulations~\cite{karimi2021algorithmic} extend these objectives to ensure that counterfactuals correspond to meaningful, achievable interventions. These diverse objectives—correctness, proximity, sparsity, plausibility, and causality—define complementary yet often competing aspects of interpretability. 

Despite their conceptual richness, most counterfactual formulations can be expressed within a common optimization framework:
\[
\mathbf{x}^* =
\arg\min_{\mathbf{x}}
\Big[
\mathcal{L}_{\text{cls}}(f(\mathbf{x}), y^*)
+ \lambda\, d(\mathbf{x}, \mathbf{x}^{(o)})
+ \Omega(\mathbf{x},\mathbf{x}^{(o)})
\Big],
\]
where $\mathcal{L}_{\text{cls}}$ enforces classification consistency, $d(\cdot,\cdot)$ penalizes deviation from the original input, and $\Omega(\mathbf{x}, \mathbf{x}^{(o)})$ regularizes the structure of the modification. In classical works such as Wachter \textit{et al.}~\cite{wachter2017counterfactual} and Dhurandhar \textit{et al.}~\cite{dhurandhar2018desiderata}, the regularizer $\Omega$ is instantiated as a simple norm, typically $L_1$ or $L_2$. Dandl \textit{et al.}~\cite{dandl2020multiobjective} later extended this paradigm to a multi-objective formulation that jointly optimizes fidelity, proximity, and sparsity, yielding a Pareto set of counterfactuals with varying trade-offs. Although effective, such methods generally assume feature independence and therefore tend to produce counterfactuals that modify isolated coordinates, ignoring correlations or spatial relations among features. This independence assumption often leads to fragmented or implausible explanations—small in magnitude but not necessarily interpretable to humans.


A parallel line of work in statistical learning has demonstrated that interpretability often improves when sparsity is replaced with \textit{structured sparsity}. Regularizers such as the group lasso~\cite{yuan2006model}, fused lasso~\cite{tibshirani2005sparsity}, and graph-guided penalties~\cite{kim2009multivariate} encourage coordinated changes among related variables, yielding solutions that align with meaningful feature groups or spatial neighborhoods. These approaches reflect an important principle: real-world data exhibit dependencies, in particular, features in tabular domains are correlated, and pixels in images form spatially coherent regions.

However, despite the maturity of structured-sparsity methods in predictive modeling, they have not been adequately exploited in counterfactual explanation frameworks. Most counterfactual methods still treat features as independent dimensions, optimizing for $L_0$-style sparsity without regard for how meaningful or coherent the resulting changes appear. Moreover, existing structured-sparsity penalties were not designed with counterfactual interpretability in mind. They promote grouped activation patterns for predictive accuracy, but they do not incorporate the semantic or domain-specific considerations required for generating human-interpretable instance-level modifications. Consequently, they cannot directly ensure that counterfactual changes are realistic and interpretable.


Beyond interpretability, robustness is a core requirement for reliable prediction systems.
Classical robustness research, most notably adversarial defenses~\cite{goodfellow2015explaining,madry2018towards} and certified smoothing~\cite{cohen2019certified}, focuses on the model’s sensitivity to small, unstructured perturbations such as $\ell_p$-bounded noise or Gaussian randomized smoothing.
These approaches are designed to capture worst-case or distributional stability, but the perturbations they consider are typically imperceptible, lack semantic meaning, and do not reflect the kinds of structured changes humans naturally interpret as meaningful modifications to the input.
Moreover, they evaluate robustness only in terms of preserving the predicted label, without characterizing how predictions shift among classes or whether alternative outputs become reachable under non-adversarial but interpretable changes.
Consequently, robustness with respect to human-understandable, structured perturbations, and the corresponding class-to-class transition behavior, remains largely unaddressed in existing robustness frameworks.

In summary, prior work has established strong foundations for counterfactual reasoning but typically optimizes for sparsity and correctness in isolation, overlooking the structural and semantic coherence of feature changes. Likewise, robustness analysis has focused on worst-case sensitivity rather than interpretable variability. Our approach aims at bridging these perspectives via an optimization framework that respects both the internal structure of the data and the external behavior of the model.

\section{Method}
\label{sec:method}

We propose a unified optimization framework for identifying \textit{minimal, interpretable, and structured modifications} that correct a misclassification while remaining close to the original instance.  
The framework jointly enforces correctness, parsimony, and realism through three complementary loss terms.

\subsection{Problem Formulation}

Let $\mathbf{x}\!\in\!\mathbb{R}^d$, $\mathbf{x} = (x_1,\dots,x_d)^t$ denote an input instance, and let 
$f:\mathbb{R}^d\!\rightarrow\!\{1,\dots,K\}$, where $K$ is the number of classes, denote a trained 
classifier that estimates posterior probabilities $p_j(\mathbf{x}) = P(y\!=\!j \mid \mathbf{x})$.  
Given a misclassified sample $\mathbf{x}^{(o)}$ with true label $y^*$, we aim to find a corrected 
input $\mathbf{x}^*$ satisfying
\begin{equation}
\label{eq:total_loss}
\mathbf{x}^*=\arg\min_{\mathbf{x}} \mathcal{L}_{\mathrm{comp}}(\mathbf{x}, \mathbf{x}^{(o)}, y^*) \; , 
\end{equation}
where the \emph{composite penalty/loss} $\mathcal{L}_{\mathrm{comp}}$ is defined as
\begin{eqnarray}
&&\mathcal{L}_{\mathrm{comp}}(\mathbf{x},  \mathbf{x}^{(o)}, y^*) 
\label{L_comp}
\\
&=&\mathcal{L}_{\mathrm{cls}}(\mathbf{x}, y^*)
+\lambda_1\mathcal{L}_{\mathrm{XA\text{-}L_0}}(\mathbf{x}, \mathbf{x}^{(o)})
+\lambda_2\mathcal{L}_{\mathrm{prox}}(\mathbf{x}, \mathbf{x}^{(o)}) \, .
\nonumber
\end{eqnarray}
The three components of the composite loss are given by
\begin{align}
\mathcal{L}_{\mathrm{cls}}(\mathbf{x}, y^*) &=
\max\!\big(0,\,\max_{j\ne y^*}(p_j(\mathbf{x})-p_{y^*}(\mathbf{x}))\big) \, ,
\label{eq:L_cls}
\\[-2pt]
\mathcal{L}_{\mathrm{prox}}(\mathbf{x}, \mathbf{x}^{(o)}) &=\|\mathbf{x}-\mathbf{x}^{(o)}\|_2^2 \, ,
\label{eq:L_prox}
\\[-2pt]
\mathcal{L}_{\mathrm{XA\text{-}L_0}}(\mathbf{x}, \mathbf{x}^{(o)})&:\text{structured explainability term} \, .
\label{eq:L_XA}
\end{align}

Here, $\mathcal{L}_{\mathrm{cls}}$ enforces the correct classification label.  
If we additionally require that the posterior of the true class exceed that of all other classes by a 
margin $\theta$, its definition becomes  
\[
\mathcal{L}_{\mathrm{cls}}(\mathbf{x}, y^*) =
\max\!\big(0,\,\max_{j\ne y^*}(p_j(\mathbf{x})-p_{y^*}(\mathbf{x}))+\theta \big),
\]
where the margin $\theta>0$ ensures that the corrected counterfactual lies safely within the 
true-class region, away from the decision boundary.  
The loss
$\mathcal{L}_{\mathrm{prox}}$ constrains the magnitude of deviation from the original sample 
$\mathbf{x}^{(o)}$, whereas $\mathcal{L}_{\mathrm{XA\text{-}L_0}}$ encourages coherent and interpretable sparsity, as elaborated below.


\subsection{Explainability-Aware \texorpdfstring{$L_{0}$ Loss (XA-$L_0$)}{L0 Loss}}
\label{sec:XA-L0}
Although the $L_0$ norm is commonly used to enforce sparsity, it ignores dependencies among features and is not differentiable, making it difficult to optimize. To capture domain-specific structure, we introduce a differentiable penalty that promotes structured activation, encouraging related features to change jointly rather than in isolation.

Define a soft activation function
\[
\sigma(x)= \frac{2}{1+\exp(-\xi x)}-1,\qquad \xi>0.
\]
In our experiments, we set $\xi=10$. Then define
\begin{eqnarray}
\label{eq:xal0}
&&\mathcal{L}_{\mathrm{XA\text{-}L_0}}(\mathbf{x}, \mathbf{x}^{(o)}) \nonumber \\
&=&\sum_{(i,j)\in\mathcal{I}}W_{i,j}\,
\sigma(|x_i-x_i^{(o)}|)\,
\sigma(|x_j-x_j^{(o)}|)\, ,
\end{eqnarray}
where $\mathcal{I}$ denotes a set of index pairs $(i,j)$ with $i\neq j$, $i, j=1, ..., d$, and $W_{i,j}\geq 0$ specifies the degree of incoherence between features $i$ and $j$, (i.e., a penalty weight that discourages their joint activation; smaller $W_{i,j}$ indicates stronger affinity and makes it less costly for the two features to change together). This construction yields a differentiable surrogate that preserves the interpretive sparsity characteristics of an $L_0$ penalty while enabling gradient-based optimization.
We also introduce several candidate formulations for $W_{i,j}$ tailored to both tabular and image data.

\subsubsection*{(a) Tabular Data: Correlation and Network-Based Grouping}

For tabular data, interpretability improves when correlated or semantically related features vary together.  
We first introduce a simple correlation-based incoherence measure:
\[
W_{i,j}
=1-\frac{|\rho_{i,j}|}{\max_{(i',j')\in\mathcal{I}}|\rho_{i',j'}|},
\]
where $\rho_{i,j}$ denotes the Pearson correlation between features $i$ and $j$.  
This construction assigns larger penalties to weakly correlated pairs and smaller penalties to strongly correlated pairs (e.g., income and credit limit, or systolic–diastolic pressure) to co-vary.
When domain knowledge is available, $W_{i,j}$ may also encode expert-specified relationships or hierarchical groupings, offering a flexible mechanism for representing application-specific dependencies.


To capture nonlinear or higher-order relationships among features, we construct a 
feature network $G=(V,E,A)$, where each feature corresponds to a node and 
edge weights encode pairwise affinities. An edge weight is defined as
$\displaystyle
A_{ij} = g(x_i, x_j) \in [0,1]$,
where $g(\cdot,\cdot)$ may represent correlation, mutual information, geometric 
proximity, or any other symmetric dependence measure.

We then apply community detection (e.g.,  spectral clustering~\cite{von2007tutorial} or modularity maximization~\cite{newman2006modularity}) 
to partition the nodes into $K_c$ groups, and let $c_i \in \{1,\dots,K_c\}$ denote the 
community label of feature $i$. Based on this partition, we define a pairwise incoherence penalty $W$ 
that controls how strongly pairs of features are encouraged to activate together:
\[
W_{i,j} =
\begin{cases}
\omega_{\mathrm{in}},  & \text{if } c_i = c_j,\\[4pt]
\omega_{\mathrm{out}}, & \text{if } c_i \neq c_j,
\end{cases}
\qquad
0 \le \omega_{\mathrm{in}} < \omega_{\mathrm{out}} \le 1.
\]
Within-community pairs receive the lower penalty $\omega_{\mathrm{in}}$, encouraging 
coherent changes among related features, whereas cross-community pairs incur the 
higher penalty $\omega_{\mathrm{out}}$, discouraging dispersed or semantically 
unrelated edits.

As a continuous alternative that does not rely on discrete communities, the incoherence measure may be defined directly from the affinity structure:
\[
W_{i,j} = 1 - \eta A_{ij}, 
\qquad \eta \in [0,1],
\]
so that strongly related features (large $A_{ij}$) receive a smaller incoherence penalty.  
This formulation unifies the similarity function $g(\cdot,\cdot)$, the graph $G$, and 
the coupling weights $W_{i,j}$ within a single coherent framework, ensuring that 
XA-$L_0$ respects the underlying feature-network geometry.


\subsubsection*{(b) Image Data: Distance- and Boundary-Based Coherence}

For images, we consider two schemes designed to encourage more interpretable modifications. The first favors changes in spatially proximate pixels over dispersed alterations across the image. The second emphasizes changes on or near edges, reflecting the intuition that edges attract greater perceptual attention and that modifying them is more likely to alter object shapes and thus influence 
classification outcomes. Correspondingly, we propose two spatial coupling strategies for defining XA-$L_0$: a pixel-distance-based scheme and an edge-focused scheme.


\paragraph*{Distance-Based Spatial Coupling}
To promote spatially localized and contiguous corrections, we couple pixels according to their spatial distance. Let the 2D coordinate of pixel $i$, $i = 1,\dots, N$ (with $N$ the total number of pixels), be denoted by $\mathbf{u}_i = (u_{i,1}, u_{i,2})$, where $u_{i,1}$ and $u_{i,2}$ represent the horizontal and vertical coordinates, respectively.  We assign each pixel pair $(i,j)$ the weight
\[
W_{i,j}
= 1 - \exp\Bigl(-\tfrac{\|\mathbf{u}_i - \mathbf{u}_j\|_2^2}{2\zeta^2}\Bigr),
\]
so that nearby pixels (small spatial separation) receive \emph{smaller} weights and distant pixels receive \emph{larger} ones. The hyperparameter $\zeta$ is set to 2.
Under the definition of $\mathcal{L}_{\mathrm{XA\text{-}L_0}}$ in Eq.~\eqref{eq:xal0}, this 
design makes it inexpensive to activate neighboring pixels jointly and costly to activate distant 
pixels, thereby encouraging compact, spatially coherent perturbation regions rather than isolated 
pixel-level changes.

\paragraph*{Edge-Focused Coherence}
For image data, a counterfactual correction should identify \emph{where} and \emph{how} the model’s perception must change in order to alter its prediction.  
Unconstrained pixel-wise optimization often produces scattered, noise-like perturbations that lack semantic structure.  
In contrast, human perception is strongly guided by \emph{edges} and \emph{object boundaries}, where salient visual changes occur—such as shifts in shape, texture transitions, or part delineations. Encouraging modifications to concentrate near these boundaries therefore leads to counterfactuals that are more interpretable and visually coherent.

To implement this principle, we introduce an \emph{edge-aware weighting} into the XA-$L_0$ penalty. Let $M_i \in [0,1]$ denote the probability that pixel $i$, $i = 1,\dots,N$, lies on an edge, obtained from the original image $\mathbf{x}^{(o)}$ using an edge detector, in particular, Holistically-
Nested Edge Detection (HED)~\cite{xie2015holistically} in our experiments. If $M_i = 0$, pixel $i$ is treated as a non-edge pixel; otherwise, it is regarded as lying on an edge.
Let $D_M(i)$ denote the Euclidean distance from pixel $i$ to the nearest detected edge. 
Pixels located on or near edges should be favored for modification, whereas pixels far inside 
homogeneous regions should be discouraged.

We encode this preference using the edge-aware function
\[
\alpha_{\mathrm{edge}}(i)
= \alpha_{\min}
  + (1-\alpha_{\min})
    \exp\!\Biggl(-\frac{D_M(i)^2}{2\tau^2}\Biggr),
\]
where $\alpha_{\min}\in(0,1)$ adjusts the minimum value for $\alpha_{\mathrm{edge}}(i)$, which decreases monotonically with $D_M(i)$. Hyperparameter $\tau$ determines how fast $\alpha_{\mathrm{edge}}(i)$ reduces when $D_M(i)$ increases. 

We also take into account the probability of pixel $i$ lying on an edge and further define 
\[
\tilde{\alpha}_{\mathrm{edge}}(i)
= \bigl(1-\kappa + \kappa\, M_i \bigr)\,\alpha_{\mathrm{edge}}(i),
\]
where $0<\kappa<1$ controls the influence of $M_i$.  

Since the XA-$L_0$ penalty operates on per-pixel activations, we convert this 
preference into a penalty weight:
\[
W^{\mathrm{ed}}_i
= \varepsilon 
  + (1-\varepsilon)\,\bigl(1 - \tilde{\alpha}_{\mathrm{edge}}(i)\bigr),
\]
with a small constant $\varepsilon>0$ ensuring nonzero gradients.

The resulting edge-aware penalty is
\[
\mathcal{L}^{\mathrm{ed}}_{\mathrm{XA\text{-}L_0}}(\mathbf{x}, \mathbf{x}^{(o)})
= \sum_{i=1}^{N} W^{\mathrm{ed}}_i\,\sigma(|x_i-x_i^{(o)}|) \, ,
\]
which encourages smooth, spatially coherent adjustments along boundaries rather 
than diffuse, noise-like pixel-level perturbations. 

\subsection{Optimization and Thresholding}
The composite loss $\mathcal{L}_{\mathrm{comp}}(\mathbf{x},\mathbf{x}^{(o)}, y^*)$ is minimized with respect to $\mathbf{x}$, 
while the classifier $f(\cdot)$ remains fixed. The optimization procedure consists of two steps:
\begin{enumerate}
\item
Initialize $\mathbf{x}$ with the original instance $\mathbf{x}^{(o)}$, and apply ADAM to solve optimization problem~\eqref{eq:total_loss}.  
Denote the resulting counterfactual sample by $\mathbf{x}^*$.
\item
Apply hard thresholding (zero-clipping) to set near-zero changes to zero, thereby removing residual 
numerical artifacts from the optimization.
In particular, suppose $t$ is the threshold and apply thresholding  individually to the value at each pixel $x^*_i$:
\[
\hat{x}^{*}_i=
\begin{cases}
x^*_i,& |x^*_i-x^{(o)}_i|>t,\\
x^{(o)}_i,& \text{otherwise} \, .
\end{cases}
\]
For tabular data, since the inputs are standardized, we set the threshold to $t = 0.05$.  For image data with pixel values in the range $[0, 255]$, we select $t \in [1, 10]$ dynamically. This choice is motivated by the fact that a modification smaller than 10 intensity levels (under the 
standard one-byte-per-channel representation) is typically imperceptible to the human eye. For $t = 1, 2, \dots, 10$, we apply each threshold successively and validate the resulting $\hat{\mathbf{x}}^*$ to ensure that the corrected sample still yields the desired classification 
label. We then set $t$ to the largest value that preserves the correct label. For image datasets in which pixel values are normalized to $[0,1]$, the thresholds are adjusted to $t = j/255$, $j = 1, \dots, 10$.
\end{enumerate}

The resulting $\hat{\mathbf{x}}^*$ constitutes a \textit{minimal sufficient correction}, namely the smallest set of changes (after zero-clipping) required to flip the classification outcome to the true class.  
Its binary mask $Z_i=I\!\left(|\hat{x}^*_i - x^{(o)}_i| > 0\right)$, where $I(\cdot)$ is the indicator function that equals 1 when the argument is true, provides key information for explaining the misclassification of the original sample.



\subsection{Quantifying Structural Coherence}

To assess the interpretability of tabular counterfactual edits, we evaluate how well the modified 
features form a compact and internally coherent group within the feature–affinity network.  
Let $\mathcal{S} = \{\, i : \hat{x}^*_i \neq x^{(o)}_i,\; 1 \le i \le d \,\}$ denote the set of 
changed features (with $d$ the input dimension), and let $W \in \mathbb{R}^{d \times d}$ be the 
pairwise incoherence penalty matrix used in the XA-$L_0$ regularizer, where smaller $W_{i,j}$ indicates stronger feature affinity and larger $W_{i,j}$ indicates weaker relation between features $i$ and $j$ (and $W_{i,i}$ takes its minimum value).

We define the \emph{structural incoherence score}:
\begin{equation}
\label{eq:phi-new}
\phi(\mathcal{S})
=\frac{1}{d\,|\mathcal{S}|}
\sum_{i \in \mathcal{S}}\sum_{j \in \mathcal{S}}
\exp\bigl(\psi\, W_{i,j}\bigr),
\end{equation}
where $\psi>0$ controls the sensitivity to pairwise affinity.

Since $W_{i,j}$ is small for closely related features, the term 
$\exp(\psi W_{i,j})$ becomes small for coherent pairs and larger for weakly 
related or independent pairs.  
Consequently, $\phi(\mathcal{S})$ remains small when $\mathcal{S}$ forms a compact cluster of 
mutually similar features and increases when the modified features are weakly related.  
Normalization by $d|\mathcal{S}|$ ensures comparability across datasets and prevents systematic 
favoring of counterfactuals that modify many features.  
Low values of $\phi(\mathcal{S})$ therefore indicate that the counterfactual corresponds to a compact, 
semantically meaningful edit aligned with the underlying structural dependencies of the dataset.

\subsection{Robustness via Tolerance-Region Confusion Matrix}
\label{sec:robustness}

To complement counterfactual explanations with a quantitative assessment of model stability, 
we evaluate how predictions behave under \emph{interpretable} perturbations rather than 
adversarial or norm-bounded noise.  
The key idea is to use the same penalty that governs counterfactual modifications, whether 
based on proximity, structured sparsity, or any domain-specific interpretability cost, to 
define a tolerance region around each instance.  
Within this region, we examine which classes become reachable and summarize the resulting 
transitions in a matrix-level representation.


The \emph{Tolerance-Region Confusion Matrix (TOR-CM)} formalizes this idea.  
For each original input $\mathbf{x}^{(o)}$, we define a tolerance region containing all 
inputs that differ from $\mathbf{x}^{(o)}$ by at most $\tau$ with respect to a chosen loss 
function.  
TOR-CM summarizes, for each true class $i$, the empirical transition probabilities to predicted classes $j$ induced by tolerance-bounded perturbations.
This yields a class-by-class characterization of prediction stability under structured, interpretable 
perturbations, rather than arbitrary norm-bounded noise.

For each sample $\mathbf{x}^{(o)}$, let $\mathcal{L}(\mathbf{x}, \mathbf{x}^{(o)})$ denote a general loss 
between $\mathbf{x}$ and $\mathbf{x}^{(o)}$. The \emph{tolerance region} is defined as:
\begin{equation}
\label{eq:tolerance-region}
\mathcal{T}_{\tau}(\mathbf{x}^{(o)})
=\big\{\mathbf{x}' : 
\mathcal{L}(\mathbf{x}',\mathbf{x}^{(o)})
\le\tau\big\},
\end{equation}
where $\mathcal{L}(\cdot,\cdot)$ may take various forms—for example 
$\|\mathbf{x}'-\mathbf{x}^{(o)}\|_2^2$ or, in our setting, the combined loss 
$\mathcal{L}_{\mathrm{XA\text{-}L_0}} + \mathcal{L}_{\mathrm{prox}}$.  

A \emph{reachable class set} at $\mathbf{x}^{(o)}$ is defined as
\[
\mathcal{R}(\mathbf{x}^{(o)})
=\big\{j:\exists\,\mathbf{x}'\!\in\!\mathcal{T}_{\tau}(\mathbf{x}^{(o)}),f(\mathbf{x}')=j\big\}.
\]
Suppose the dataset is partitioned into $\mathcal{D}_i$, $i=1,\dots,K$, where 
$\mathcal{D}_i$ contains all samples with true class label $i$.  
The TOR-CM $C = (C_{i,j})_{i,j=1,\dots,K}$ is then defined by
\begin{equation}
\label{eq:tor-cm}
C_{i,j}
=\!\!\sum_{\mathbf{x}^{(o)}\in\mathcal{D}_i} I(j\in \mathcal{R}(\mathbf{x}^{(o)})) \, , \;\; i, j=1, ..., K \; .
\end{equation}
Row-normalizing by $|\mathcal{D}_i|$ yields $\tilde C_{i,j}=C_{i,j}/|\mathcal{D}_i|$, which can be interpreted as an empirical estimate of the probability that a class-$i$ instance can be perturbed into class $j$ within tolerance $\tau$. The diagonal elements $\tilde C_{i,i}$ reflect classifier robustness, while the off-diagonal entries quantify feasible class transitions under structured perturbations.

Finally, we define two summary metrics, capturing robust accuracy and vulnerability respectively. Let $N=\sum_{i=1}^K|\mathcal{D}_i|$ denote the total number of samples. We define
\begin{equation}
\label{eq:robust-metrics-accuracy}
\gamma_{\mathrm{a}}
=\frac{1}{N}\sum_{i=1}^K C_{i,i} 
\end{equation}

\begin{equation}
\label{eq:robust-metrics-vul}
\gamma_{\mathrm{v}}
=\frac{1}{K N}\sum_{i=1}^{K}\sum_{j: j\neq i} C_{i,j} \;\;.
\end{equation}



\section{Experiments}
\label{sec:experiments}
In this section, we evaluate our method on both tabular and image data and conduct a robustness analysis using the proposed TOR-CM. A central goal of the evaluation is to determine whether XA-$L_0$ produces compact, structurally coherent modifications. Through TOR-CM, we further investigate how such human-interpretable perturbations reveal aspects of model robustness that remain invisible under conventional adversarial or norm-bounded evaluations. Across all experiments comparing XA-$L_0$ with existing methods, we keep the classifier architecture, optimization budget, thresholding rules, and training procedures fixed to ensure that any observed differences in sparsity, coherence, or robustness stem from the loss formulations themselves rather than implementation details.

\subsection{Tabular Data}
\label{sec:tabular}

\begin{figure*}[!t]
\centering
\subfloat[Breast Cancer: incoherence $\phi(\mathcal{S})$]
  {\includegraphics[width=0.47\linewidth]{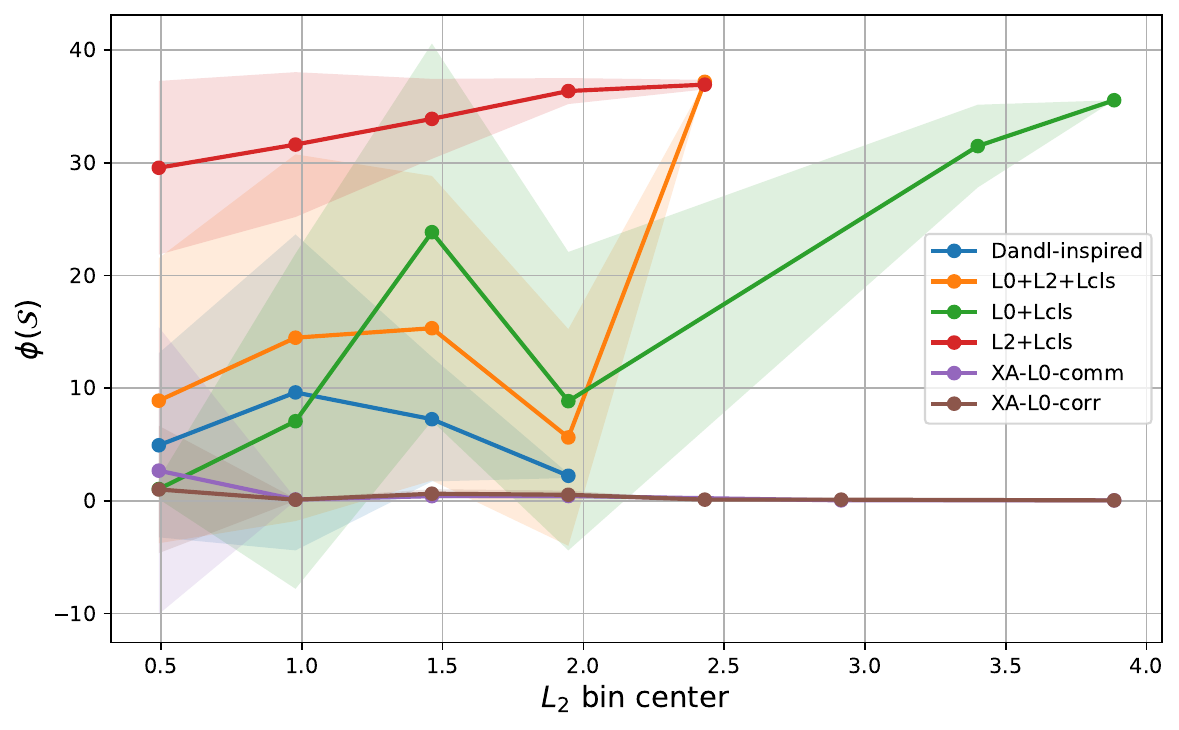}}
\hfill
\subfloat[Breast Cancer: sparsity $n$]
{\includegraphics[width=0.47\linewidth]{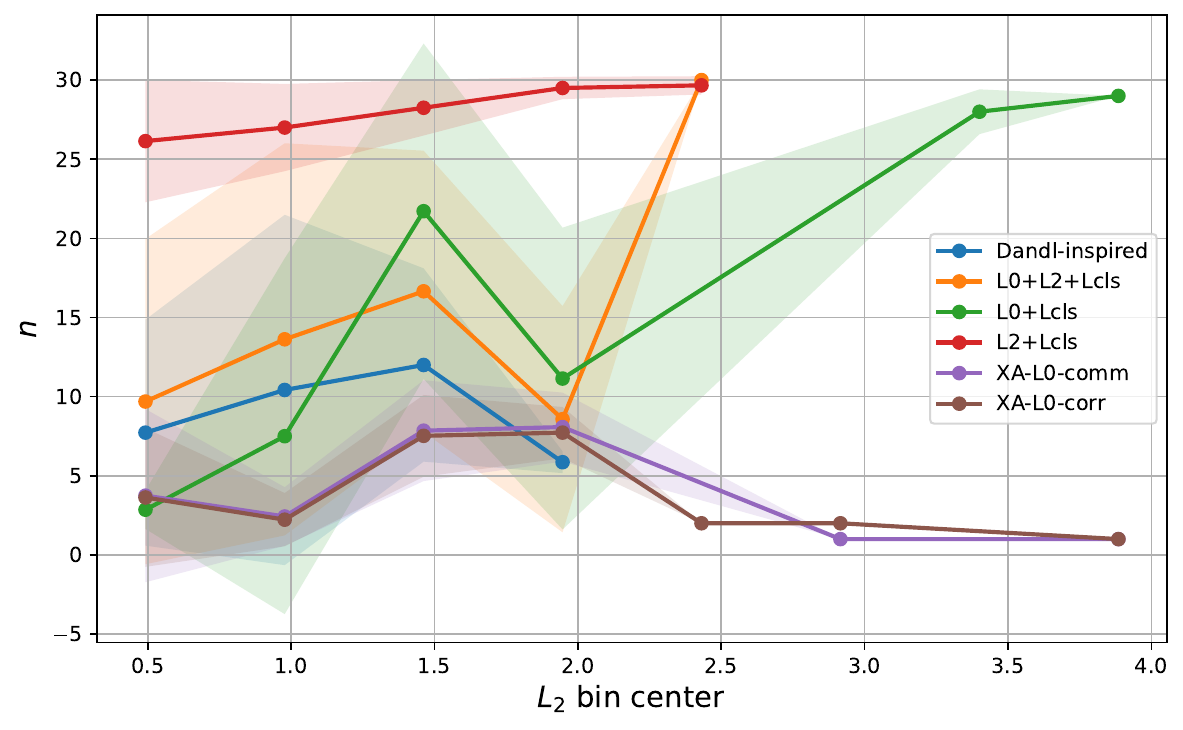}}
\hfill
\subfloat[Coil2000: incoherence $\phi(\mathcal{S})$]
    {\includegraphics[width=0.47\linewidth]{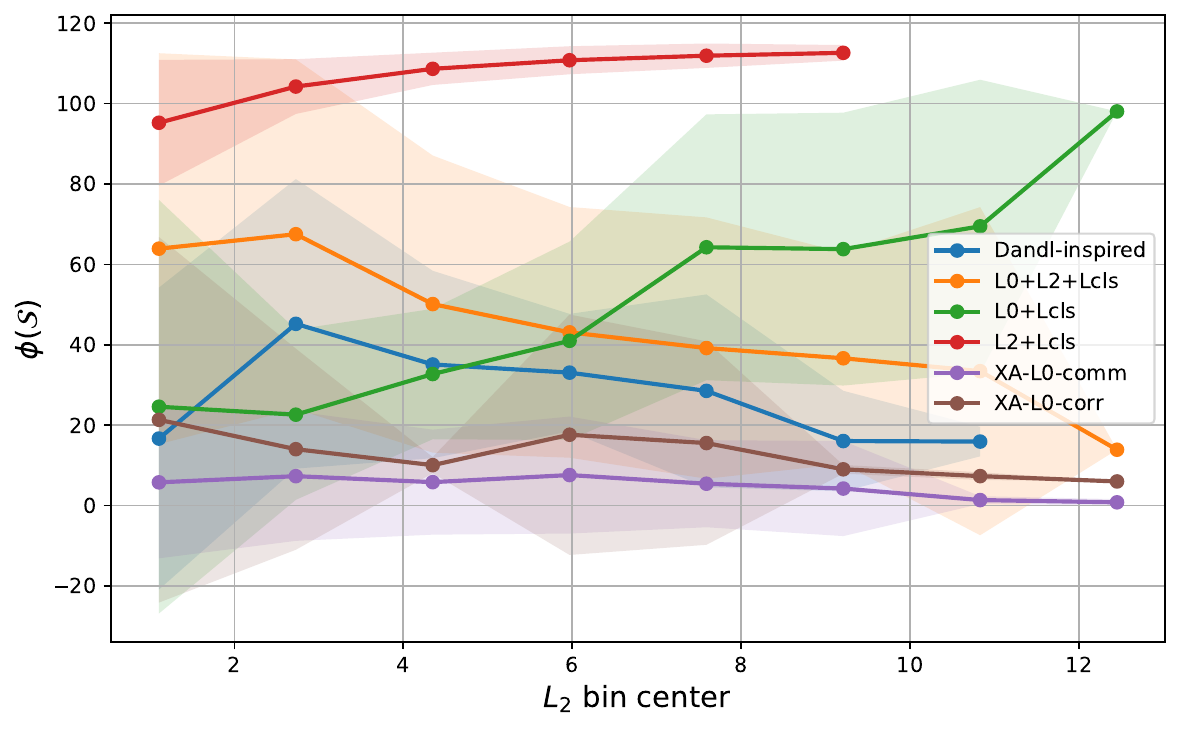}
    \label{fig:d_phi_binned}}
    \hfill
\subfloat[Coil2000: sparsity $n$]
  {\includegraphics[width=0.47\linewidth]{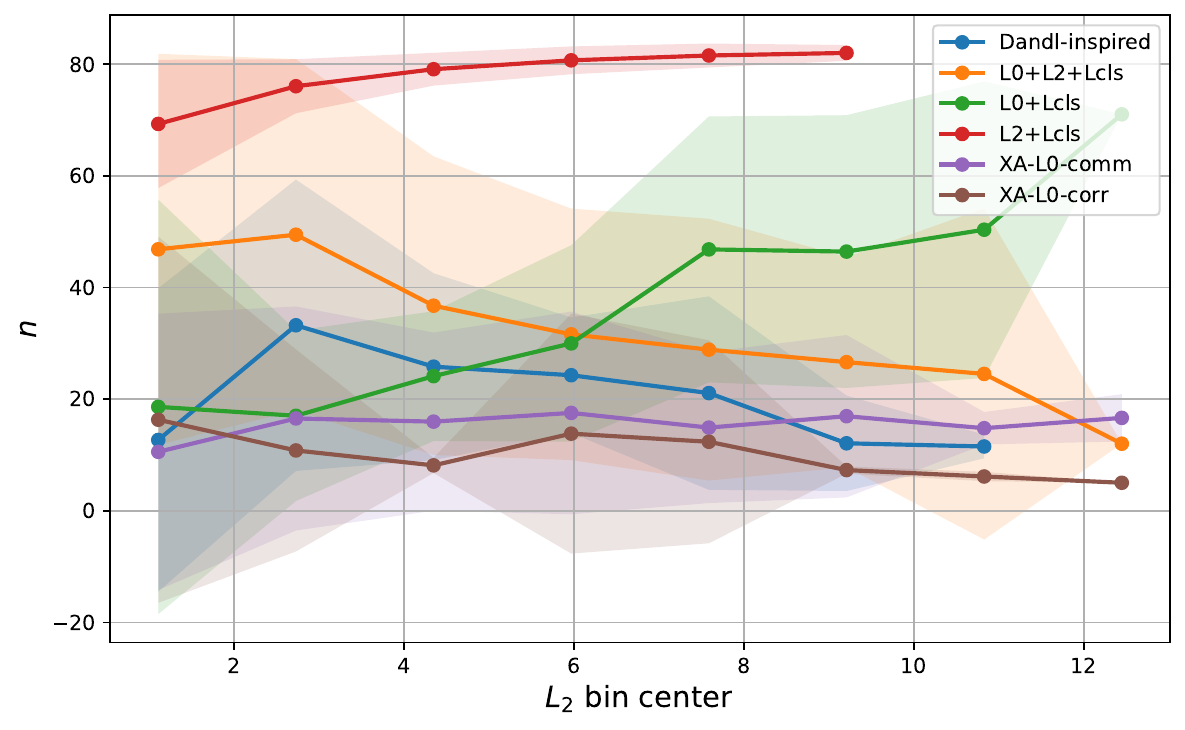}
  \label{fig:d_s_binned}}
\caption{Trade-offs between proximity ($L_2$), structural incoherence ($\phi$), and sparsity ($n$) for the Breast Cancer and Coil2000 datasets.
Left: structural incoherence versus $L_2$.
Right: sparsity versus $L_2$.
For clarity of visualization, instances are binned according to their $L_2$ values, and the average of each metric within each bin is reported.
Solid lines denote the mean value in each bin, while shaded bands indicate one standard deviation.
}
\label{fig:tabular_binned_phi_s}
\end{figure*}

We evaluate XA-\(L_0\) on a diverse collection of seven tabular datasets spanning binary and multiclass classification. Table~\ref{tab:datasets} provides the basic information about the datasets. 
For each dataset, we standardize the features, create training--test partitions, and train a two-layer MLP classifier to obtain consistently strong predictive performance. Counterfactuals are generated only for misclassified test samples, and all methods use identical optimization budgets, step sizes, and post-hoc zero-clipping rules to ensure a fair comparison.

\begin{table}[h]
\centering
\renewcommand{\arraystretch}{1.15}
\setlength{\tabcolsep}{5pt}
\begin{tabular}{lcccl}
\hline
Id & Dataset & Dimension & Number of classes \\
\hline
1 & Wine                     & 13 & 3   \\
2 & Breast Cancer (WDBC)     & 30 & 2   \\
3 & Iris                     & 4  & 3   \\
4 & Digits (8$\times$8)      & 64 & 10  \\
5 & Wine Quality (Red)       & 11 & 6   \\
6 & Phoneme                  & 5  & 2   \\
7 & Coil2000                 & 85 & 2   \\
\hline
\end{tabular}
\vspace{2mm}
\caption{Datasets used in the tabular experiments. All features are numeric and standardized per dataset.}
\label{tab:datasets}
\end{table}

To isolate the role of structured explainability, we compare against several baselines:
\begin{enumerate}
\item
\emph{XA-$L_0$}: Our proposed method, using the complete loss function
$\mathcal{L}_{\mathrm{comp}}$ in Eq.~\ref{L_comp}. Furthermore, we experiment with two schemes for XA-$L_0$ using correlation and feature community detection respectively (see Section~\ref{sec:XA-L0}), which we refer to as XA-$L_0$-corr and XA-$L_0$-comm.
\item
\emph{Standard $L_0$}: Modify $\mathcal{L}_{\mathrm{comp}}$ by replacing $\mathcal{L}_{\mathrm{XA\text{-}L_0}}$ with the smooth $L_0$ surrogate. This method is referred to as $L_0+L_2+L_{\mathrm{cls}}$. 
\item
\emph{Baseline $L_2$}: Modify $\mathcal{L}_{\mathrm{comp}}$ by removing the term
$\mathcal{L}_{\mathrm{XA\text{-}L_0}}$. This method is referred to as $L_2+L_{\mathrm{cls}}$. 
\item
\emph{Baseline $L_0$}: Modify $\mathcal{L}_{\mathrm{comp}}$ by removing the term
$\mathcal{L}_{\mathrm{prox}}$ and replacing $\mathcal{L}_{\mathrm{XA\text{-}L_0}}$ with the plain $L_0$ norm. This method is referred to as $L_0+L_{\mathrm{cls}}$. 
\item
\emph{Dandl-inspired scalarization sweep}: A baseline inspired by Dandl et al.~\cite{dandl2020multiobjective} and their trade-off view of counterfactual generation. We obtain multiple candidate counterfactuals by sweeping scalarization weights in a gradient-based objective that combines classification, sparsity, and proximity terms.
\end{enumerate}

For the XA-$L_0$-comm variant, we construct a feature–feature affinity matrix from the training set using the absolute Pearson correlation matrix, normalized to $[0,1]$. We then apply spectral clustering with this precomputed affinity to partition the $d$ features into $K_c$ communities, where $K_c=\max\{2,\min(d-1,\ \mathrm{round}(\sqrt{d}))\}$. The resulting discrete community labels $c_i\in\{1,\ldots,K_c\}$ define a block-structured penalty matrix $W$, assigning a smaller within-community penalty $\omega_{\mathrm{in}}$ and a larger cross-community penalty $\omega_{\mathrm{out}}$ (with $W_{ii}=0$), encouraging coordinated edits within coherent feature groups.
Across all methods we report proximity ($L_2$), sparsity (that is, the number of changed features after thresholding, denoted by $n$), and the structural incoherence score $\phi(\mathcal{S})$ computed on the same dataset-level computed using the same pairwise incoherence matrix $W$. To sweep out the trade-offs in Figure~\ref{fig:tabular_binned_phi_s}, we generate multiple counterfactual solutions per test instance by varying the regularization weights in the optimization objective. For XA-$L_0$ (both corr and comm variants), we evaluate a grid over ($\lambda_{\mathrm{1}}$, $\lambda_{\mathrm{2}}$); for the baselines we sweep the corresponding penalty weights (only $\lambda_{\mathrm{2}}$ for $L_2+L_{\mathrm{cls}}$, only $\lambda_{1}$ for $L_0+L_{\mathrm{cls}}$, and ($\lambda_{1}$, $\lambda_{\mathrm{2}}$) for $L_0+L_2+L_{\mathrm{cls}}$). The resulting runs are binned by their achieved $L_2$, and we report the mean and one-standard-deviation bands of $\phi(\mathcal S)$ (or $n$) within each bin.

Figure~\ref{fig:tabular_binned_phi_s} visualizes trade-offs between proximity (measured by $L_2$ norm) and interpretability measures, specifically, the structural incoherence and sparsity based on two datasets (Breast Cancer and Coil2000). The instances are binned based on their $L_2$ values. Then the average $\phi$, $n$, and $L_2$ are computed for instances in each bin. The left column reports the average structural incoherence $\phi(\mathcal{S})$ and the right column reports the average sparsity $n$, both versus the average binned $L_2$ values. Across bins and across both datasets, XA-$L_0$ (correlation- and community-based variants) consistently yields substantially smaller $\phi(\mathcal{S})$ than the $L_2+L_{\mathrm{cls}}$ baseline, indicating that the modified features form a more compact and internally consistent group under the feature-dependence structure. This advantage is especially pronounced on Coil2000, where the $L_2+L_{\mathrm{cls}}$ baseline exhibits high incoherence penalties and large numbers of changed features, whereas XA-$L_0$ maintains low $\phi(\mathcal{S})$ while keeping $n$ small. Furthermore, XA-$L_0$-corr achieves the lowest values of both $\phi$ and sparsity $n$ across all $L_2$ values, outperforming all other methods. XA-$L_0$-comm exhibits similar performance.

\begin{table*}[t]
\centering
\renewcommand{\arraystretch}{1.15}
\setlength{\tabcolsep}{9pt}

\begin{tabular}{l|cc|cc|cc|cc|cc|cc}
\hline
& \multicolumn{2}{c|}{XA-$L_0$-corr} 
& \multicolumn{2}{c|}{XA-$L_0$-comm}
& \multicolumn{2}{c|}{$L_0$+$L_\mathrm{cls}$}
& \multicolumn{2}{c|}{$L_0$+$L_2$+$L_\mathrm{cls}$}
& \multicolumn{2}{c|}{$L_2$+$L_\mathrm{cls}$}
& \multicolumn{2}{c}{Dandl-inspired} \\
Dataset 
& $\phi$ & $n$ 
& $\phi$ & $n$ 
& $\phi$ & $n$ 
& $\phi$ & $n$ 
& $\phi$ & $n$ 
& $\phi$ & $n$ \\
\hline
Wine              & 1.162 & 1.6 & 3.138 & 1.8 & 7.713  & 3.5  & 12.615 & 5.0  & 35.381  & 12.2 & 3.678 & 2.5 \\
Breast Cancer     & 0.624 & 7.5 & 0.410 & 7.8 & 23.831 & 21.7 & 15.311 & 16.7 & 33.887  & 28.2 & 7.236 & 12.0 \\
Iris              & 0.500 & 2.0 & 0.250 & 1.0 & 0.250  & 1.0  & 0.250  & 1.0  & 1.015   & 3.0  & 0.250 & 1.0 \\
Digits            & 1.534 & 2.7 & 4.315 & 5.4 & 39.575 & 29.8 & 28.037 & 21.8 & 75.868  & 55.3 & 3.494 & 4.1 \\
Wine Quality Red  & 21.134 & 5.5 & 21.151 & 5.5 & 32.570 & 7.0  & 32.406 & 7.3  & 49.504  & 10.5 & 24.674 & 5.9 \\
Phoneme           & 9.345 & 2.4 & 20.351 & 2.6 & 36.267 & 4.1  & 27.908 & 3.2  & 41.476  & 4.8  & 21.626 & 2.0 \\
Coil2000          & 17.600 & 13.8 & 7.567 & 17.5 & 40.976 & 30.0 & 43.083 & 31.6 & 110.799 & 80.7 & 33.060 & 24.3 \\
\hline
\end{tabular}
\vspace{2mm}
\caption{Median-bin summary: for each dataset, the average $\phi$ and $n$ computed over instances in the bin corresponding to the median $L_2$ value.}
\label{tab:tabular_median_bin_summary}
\end{table*}
Results for all seven datasets are summarized in Table~\ref{tab:tabular_median_bin_summary}. 
For each dataset and method, we focus on instances that belong to the bin with the median $L_2$ value. 
We then compute the average structural incoherence $\phi$ and sparsity $n$ over these instances. 
Two consistent trends emerge. 
First, the $L_2+L_{\mathrm{cls}}$ baseline typically achieves class flips by spreading small adjustments across many features, resulting in both high incoherence and large support sizes (e.g., Digits: $\phi=75.868$, $n=55.3$; Breast Cancer: $\phi=33.887$, $n=28.2$). 
Second, introducing sparsity via the $L_0+L_2+L_{\mathrm{cls}}$ baseline improves sparsity but does not guarantee better structural incoherence, indicating that the selected features are not necessarily affinity-consistent.

XA-$L_0$ addresses this gap by explicitly coupling feature activations. On datasets with clear dependency structure, most notably Breast Cancer and Digits, XA-$L_0$-corr/comm achieves very small incoherence penalties while keeping $n$ in the single digits, demonstrating that the correction can be both compact and structurally aligned. On simpler problems such as Iris, many methods already admit near-minimal edits; accordingly, XA-$L_0$-comm and Dandl reach $\phi \approx 0.250$ with $n \approx 1$, and the difference between methods is negligible. On Wine, XA-$L_0$-corr and Dandl again provide low-$\phi$ solutions with small $n$, while the $L_2+L_{\mathrm{cls}}$ baseline remains substantially more diffuse.

The advantage of XA-$L_0$ over other methods is less pronounced on datasets with weak or relatively uniform correlations between features, such as Wine Quality Red and Phoneme. 
For these datasets, structural incoherence values are higher across all methods, suggesting that correlation-derived affinity is less informative or that meaningful dependencies are distributed across multiple predictors. 
Even in this regime, XA-$L_0$ remains competitive in terms of sparsity (with single-digit $n$) and avoids the extreme diffuseness exhibited by the $L_2+L_{\mathrm{cls}}$ baseline. 
Overall, the table supports the central claim: whenever the data exhibit an interpretable dependency structure, incorporating this structure into the sparsity regularizer yields counterfactual corrections that are simultaneously more coherent and more parsimonious than proximity-only or unstructured-sparsity alternatives.

To illustrate how the method quantifies a decision’s proximity to changing, we consider a test applicant from the COIL2000 insurance dataset who is predicted as not likely to purchase (class 0) with purchase probability $0.430$. Using the training-set correlation structure, we partition the $d=85$ features into $K_c=9$ communities and then compute targeted counterfactuals under the prescribed tolerance and thresholding rule. For XA-$L_0$-comm, the purchase probability increases from $0.430$ to $0.500$ with only $|S|=3$ meaningful feature edits, touching 3 communities (one feature each from communities 0, 2, and 6). The dominant change is in the relationship-status indicator (MRELOV), accompanied by smaller shifts in a brand-related insurance attribute (ABRAND) and the trailer-policy contribution variable (PAANHANG). In comparison, a Dandl baseline reaches a similar probability threshold ($0.500$) but requires a less compact change: $|S|=8$ feature edits spanning 5 communities, with prominent adjustments including social-class indicators (e.g., MOPLHOOG, MAUT0), relationship status (MRELOV), and several insurance product variables (ABRAND, PBYSTAND, PAANHANG), along with additional attributes (AWERKT, AZEILPL). From an application perspective, such results provide actionable insight into decision sensitivity: they reveal whether a borderline “not likely to purchase” prediction is driven by a small set of coherent factors (as in the XA-$L_0$-comm explanation) or whether crossing the decision boundary would require broader, multi-factor shifts. This distinction can support model debugging and policy design, for example, by flagging cases where the decision hinges on a single sensitive attribute, or by identifying which groups of features most strongly control the transition from non-purchase to purchase predictions.


\subsection{Image Data}
\label{subsec:image-data}
We evaluate XA-$L_0$ on two image benchmarks with different levels of visual complexity.
MNIST offers a controlled grayscale setting where counterfactual corrections often correspond to small, spatially localized stroke edits. We use penalty $\mathcal{L}_{\mathrm{XA\text{-}L_0}}$ on MNIST. Flowers-102 consists of natural RGB photographs, where successful corrections can involve more complex appearance changes. On this dataset, we experiment with the edge-aware penalty $\mathcal{L}^{\mathrm{ed}}_{\mathrm{XA\text{-}L_0}}$ and test whether it encourages modifications to concentrate near edges, which are more visually salient than smooth regions.
In both experiments we start from misclassified test instances and generate \emph{targeted}
counterfactuals toward the ground-truth label while keeping the classifier fixed.
After optimization, we apply zero-clipping as described in Sec.~\ref{sec:method} to remove small-magnitude
artifacts.

\begin{figure*}[!t]
\centering
\subfloat[XA-$L_0$]{\includegraphics[width=0.32\linewidth]{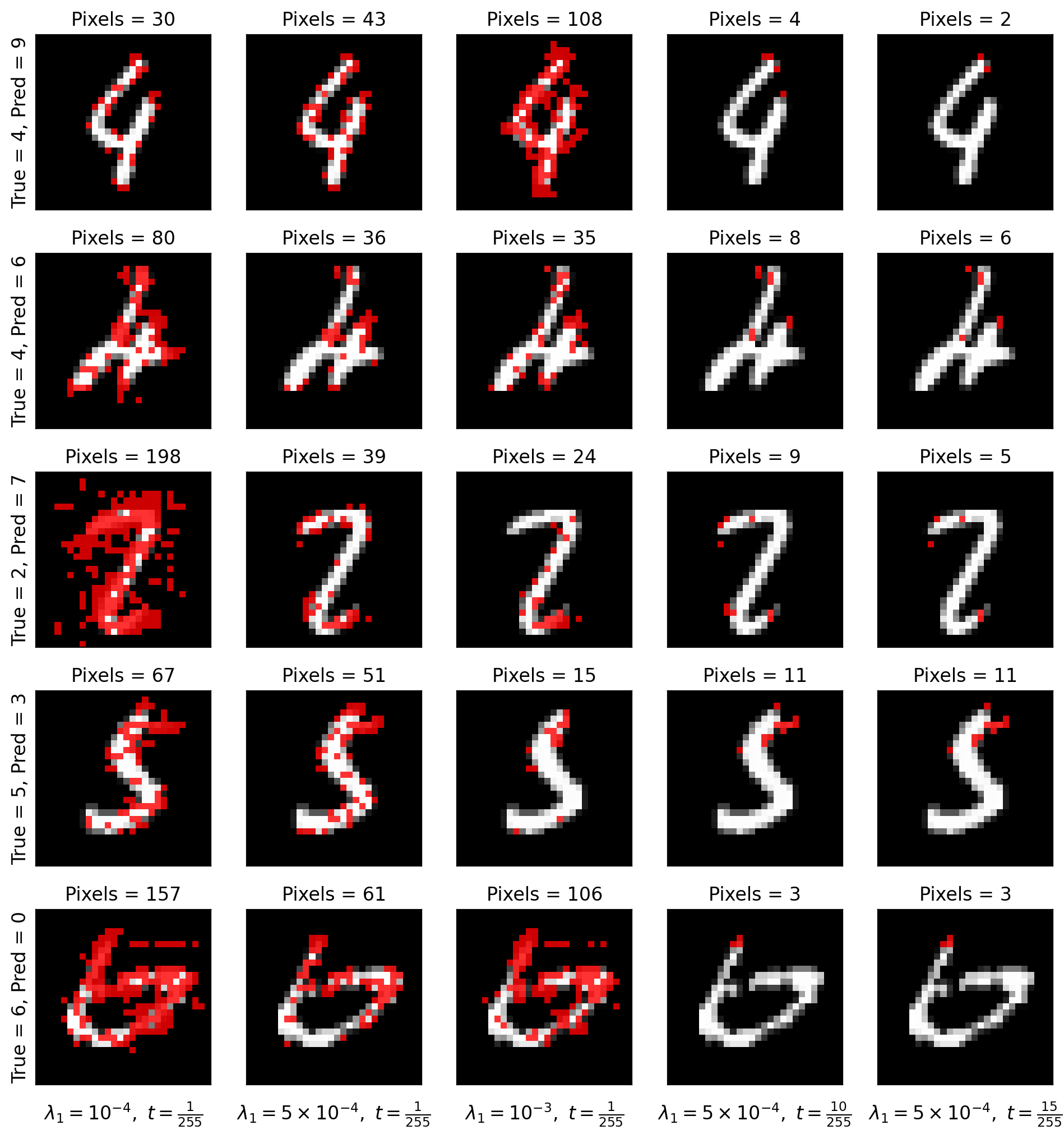}\label{fig:mnist_xal0}}
\hfill
\subfloat[$L_0$]{\includegraphics[width=0.32\linewidth]{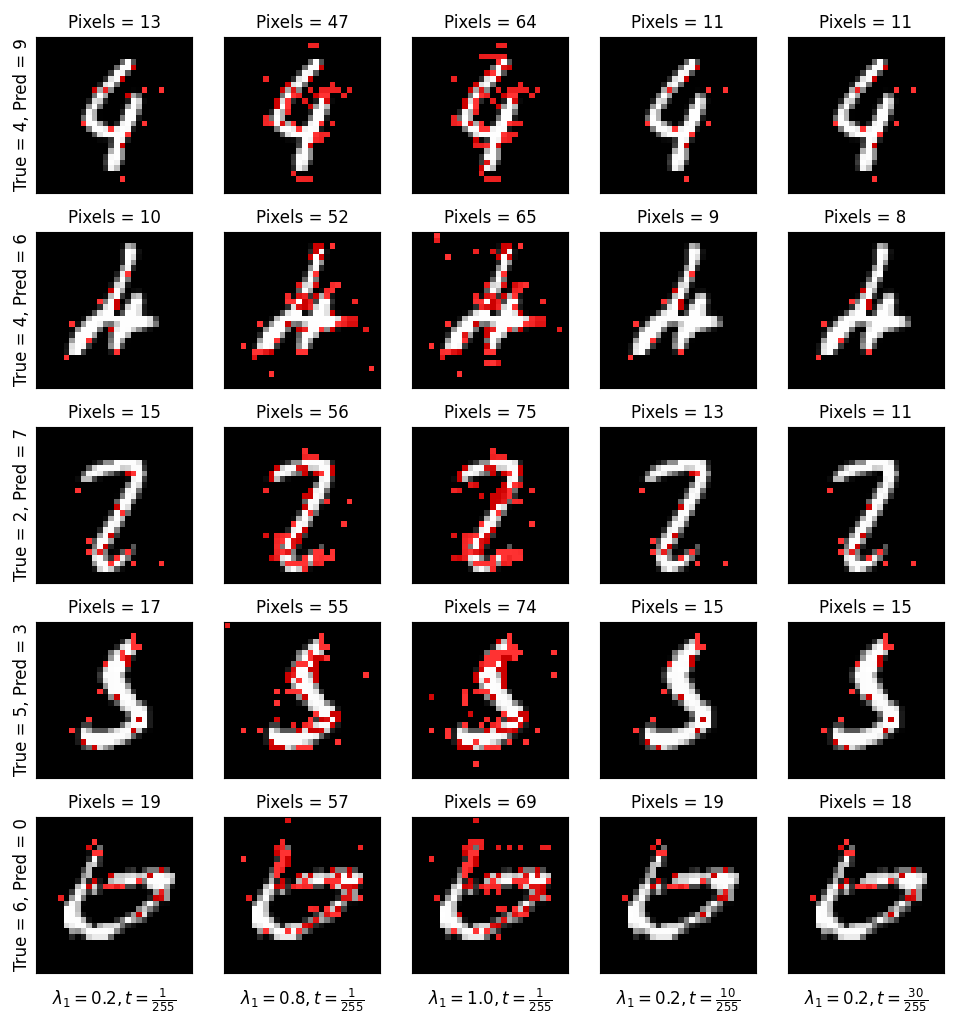}\label{fig:mnist_l0}}
\hfill
\subfloat[CEMAE]{\includegraphics[width=0.32\linewidth]{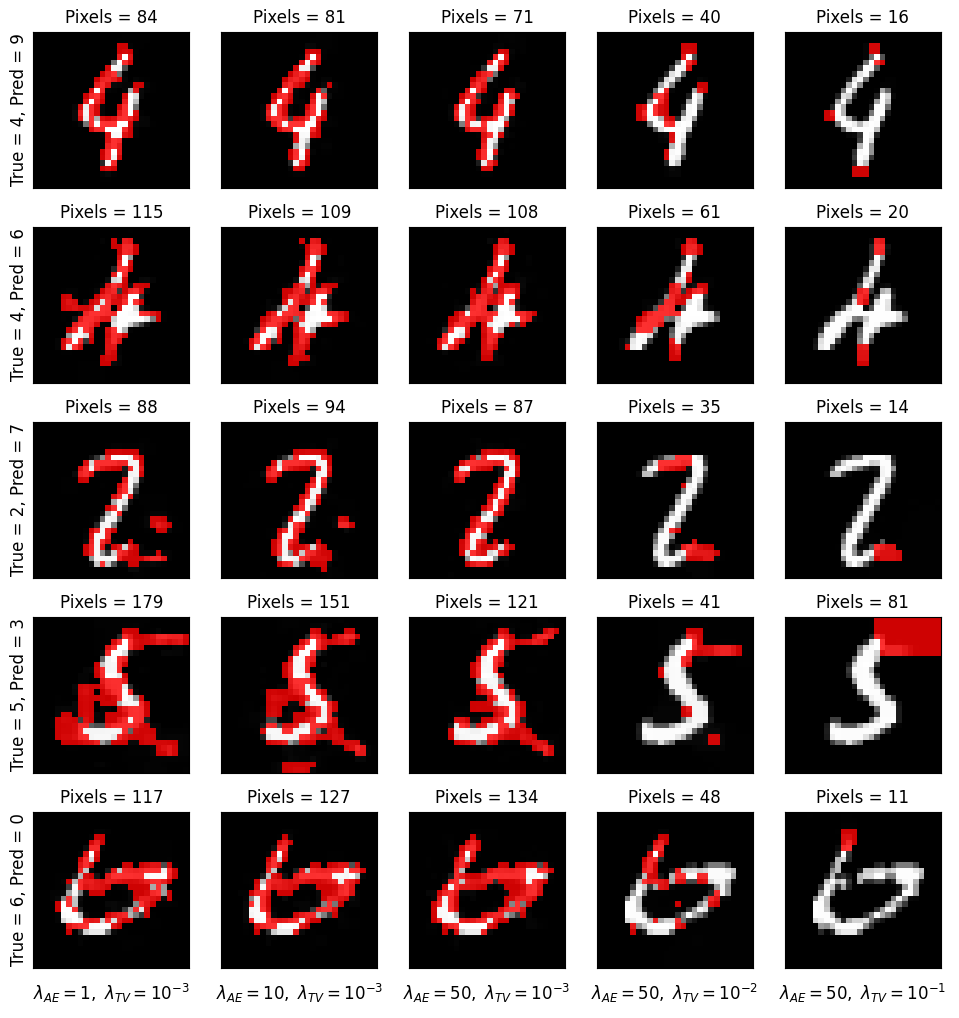}\label{fig:mnist_cemae}}
\caption{
Targeted MNIST corrections on five misclassified test images.
Red overlay indicates pixels that changed. XA-$L_0$ produces compact, stroke-aligned edits across a range of $(\lambda_1,t)$ settings, while $L_0$ and CEMAE tend to yield less coherent or more diffuse changes.
}
\label{fig:mnist}
\end{figure*}
\subsubsection*{(a) MNIST: distance-based incoherence and sparsity control}

A pretrained CNN classifier (two $3\times 3$ convolution blocks with ReLU and $2\times 2$ max pooling,
followed by two fully connected layers) is used and kept fixed during counterfactual generation.
We show results for five misclassified test images and compare three methods: XA-$L_0$,
a generic sparsity-driven $L_0$ baseline (referred to as $L_0$), and \emph{CEMAE}~\cite{dhurandhar2018explanations} (a contrastive explanation method with an
autoencoder prior).  For XA-$L_0$, we fix $\lambda_2=0.1$ and vary the sparsity weight
$\lambda_1$ and clipping threshold $t$ over
\[
(\lambda_{1},t)\in\Bigl\{
\begin{aligned}
&(10^{-4},\tfrac{1}{255}),\,
(5\times 10^{-4},\tfrac{1}{255}),\,
(10^{-3},\tfrac{1}{255}),\\
&(5\times 10^{-4},\tfrac{10}{255}),\,
(5\times 10^{-4},\tfrac{15}{255})
\end{aligned}
\Bigr\}.
\]
Figure~\ref{fig:mnist} presents the resulting counterfactual images. The red overlay marks the set of pixels changed from the original image, directly visualizing the number of modifications required to reach correct classification. Two consistent trends appear. 
First, as shown by columns 1--3 with a fixed $t=\tfrac{1}{255}$, increasing $\lambda_1$  produces progressively sparser
solutions: the red overlay shrinks to a smaller number of pixels.  Second, for fixed $\lambda_1=5\times 10^{-4}$ (shown by columns 1,4,5), increasing $t$ 
removes speckles caused by low-magnitude changes and preserves only the dominant contiguous changes that drive the
correction.  As expected, $\lambda_1$ plays the primary role of controlling how many pixels are changed, while thresholding at $t$ can further reduce the number of pixel changes required to steer the classification to the target class.
Compared to the two baseline methods, XA-$L_0$
consistently produces spatially compact edits concentrated on or near digit strokes rather than in the black background, whereas $L_0$ and CEMAE tend to yield less coherent selections or more extensive modifications.  Overall, the distance-based coupling produces minimal corrections that are both sparse and visually meaningful.

\subsubsection*{(b) Flowers-102: edge-focused incoherence on natural images}
Next, we evaluate XA-$L_0$ in a natural-image setting using Flowers-102 dataset.
We fine-tune a ResNet-18 classifier for 102-way classification and keep it fixed during
counterfactual generation.  Here we use the edge-focused coherence penalty $\mathcal{L}^{\mathrm{ed}}_{\mathrm{XA\text{-}L_0}}$,
which favors changing pixels near detected edges and discourages edits deep inside homogeneous
regions. To threshold pixel changes after minimizing the penalty function, we normalize pixel values to $[0, 1]$ and select the largest $t^\star$ from the set
$\{1,\ldots,10\}/255$ such that the target label is still obtained after
zero-clipping. 

\begin{figure}[h]
\centering
\includegraphics[width=0.95\linewidth]{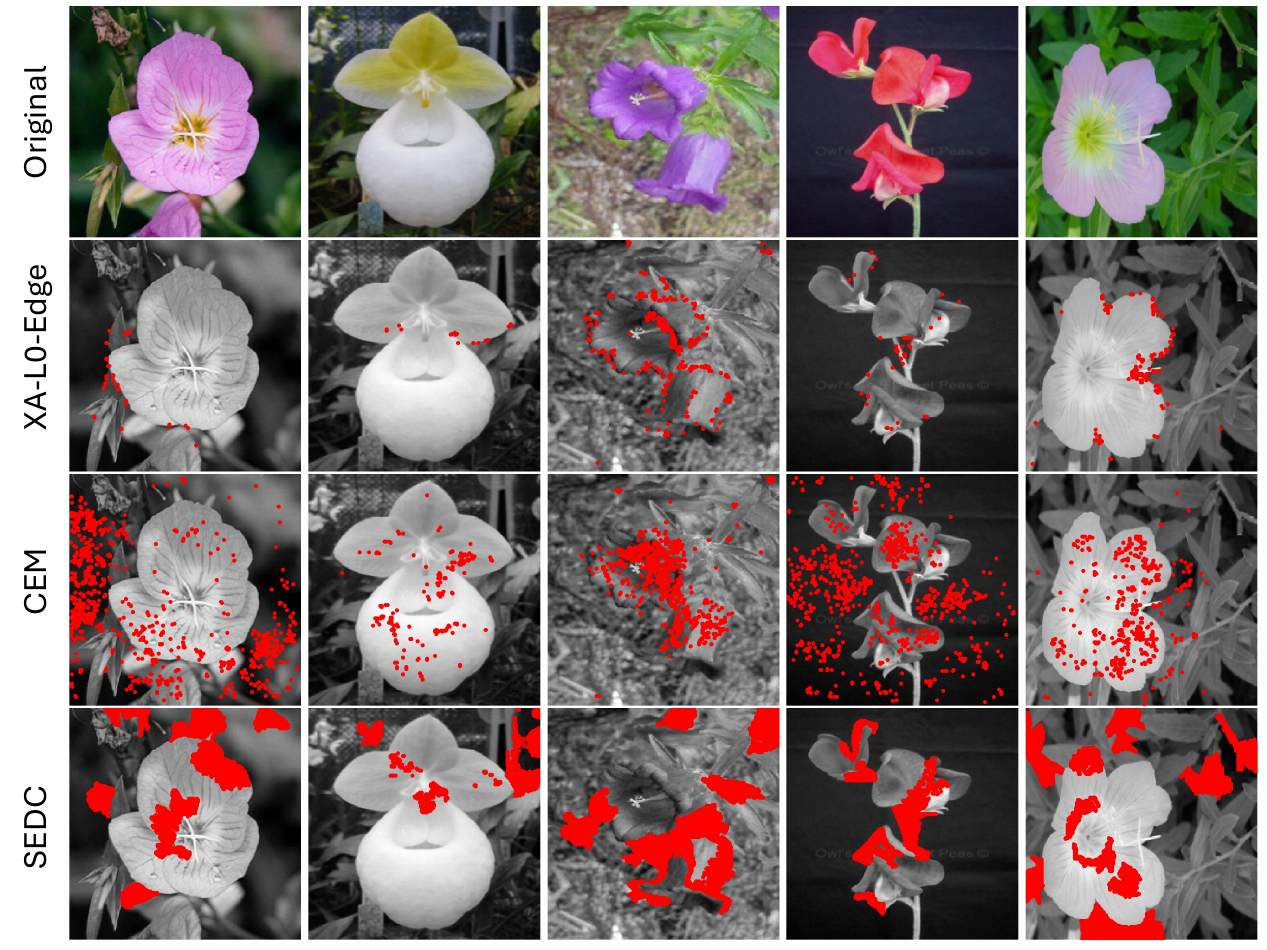}
\caption{
Flowers-102 targeted corrections on five misclassified test images.
Rows show the original image and counterfactuals for XA-$L_0$ (edge-focused), CEM, and SEDC.
}
\label{fig:flowers_grid}
\end{figure}

We compare against two representative baseline counterfactual methods for images:
\emph{CEM}~\cite{dhurandhar2018explanations}, an optimization-based method with elastic-net regularization but no explicit spatial or
edge coherence, and \emph{SEDC}~\cite{vermeire2022explainable}, a superpixel-based evidence counterfactual method that achieves
the target by replacing selected segments.  

\begin{figure}[!h]
\centering
\includegraphics[width=\linewidth]{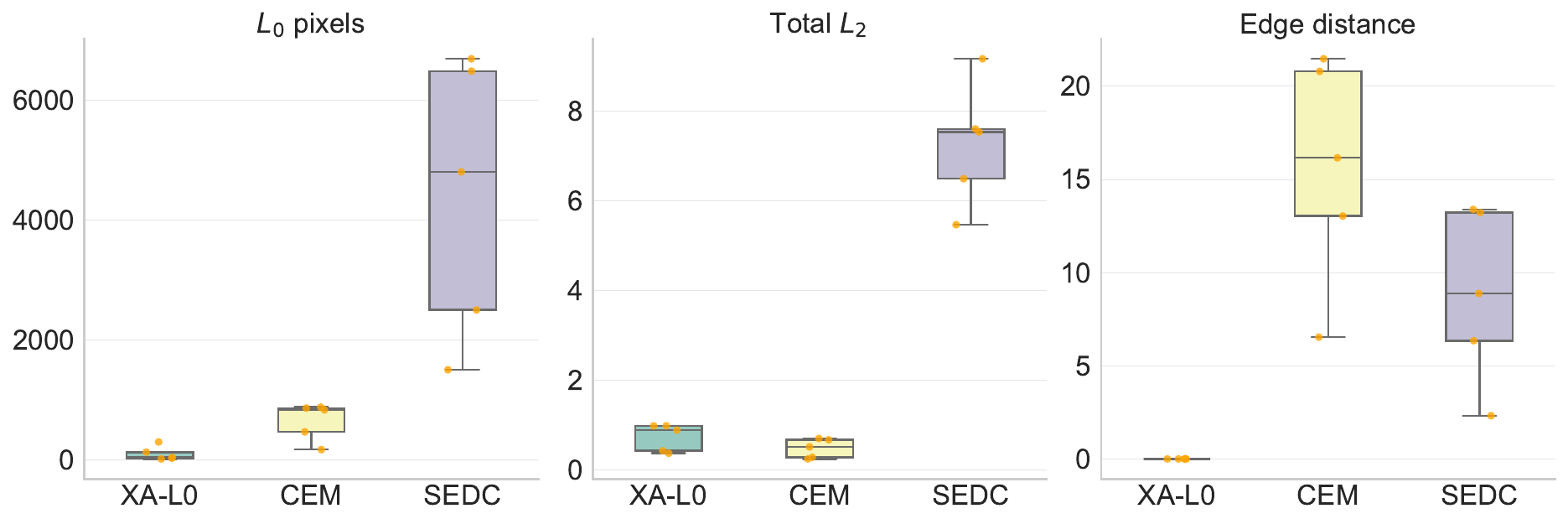}
\caption{
Comparison of three methods, XA-$L_0$, CEM, and SEDC, based on 50 misclassified Flowers-102 images using boxplots. (a) Number of changed
pixels ($L_0$ mask count), (b) The total $L_2$ distance, (c) The mean distance of changed pixels to the nearest detected edge.
}
\label{fig:flowers_metrics}
\end{figure}

Example counterfactual images for five misclassified images are shown in Figure~\ref{fig:flowers_grid}, with the true class as the target. Again, the red overlays indicate the set of pixels whose values are changed. XA-$L_0$ typically produces a small number of visually coherent modifications concentrated near object boundaries and salient contours, whereas CEM often requires more dispersed pixel changes and SEDC produces larger region-level edits that reflect its segment-replacement mechanism. These qualitative observations are consistent with the quantitative comparison in Figure~\ref{fig:flowers_metrics} over 50 images. The figure shows boxplots of the $L_0$ mask count (i.e., the number of changed pixels after zero-clipping), the $L_2$ penalty (the total $L_2$ distance over all changed pixels), and the mean distance from each changed pixel to its nearest edge pixel in the image plane. XA-$L_0$ achieves substantially smaller $L_0$ values (i.e., fewer changed pixels) and smaller $L_2$ values than SEDC. Compared with CEM, XA-$L_0$ attains similar $L_2$ values but considerably smaller $L_0$ values. Moreover, XA-$L_0$ yields a significantly smaller mean distance to edges than the other methods. Overall, results on Flowers-102 demonstrate that incorporating edge-aware coherence produces counterfactual corrections that are both parsimonious and aligned with human-perceptual structure.

\subsection{Robustness Assessment with TOR-Confusion Matrix}

To assess robustness under bounded, interpretable modifications, we use the Tolerance-Region Confusion Matrix (TOR-CM). 
Whereas a standard confusion matrix evaluates predictions only at the original inputs, TOR-CM characterizes the probability of class-to-class transitions when inputs are allowed to vary within a prescribed tolerance region.

\begin{figure}[h]
  \centering
  \includegraphics[width=\linewidth]{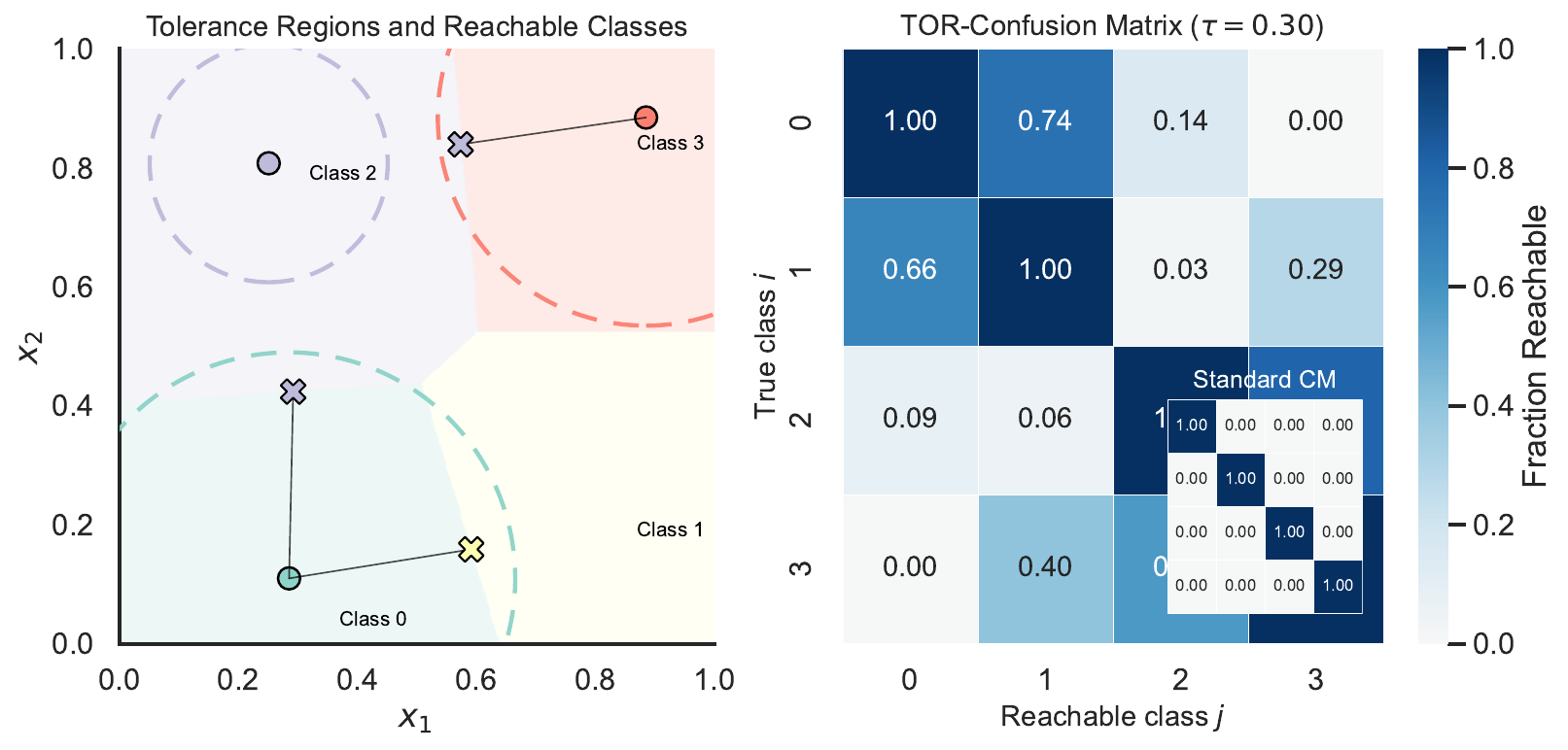}
  \caption{Illustration of TOR-CM on a 2D four-class toy problem. Left: tolerance regions (dashed circles) and examples of feasible points (crosses) inside the region that achieve specified target classes. Right: TOR-CM at shared radius $\tau=0.30$, where each entry is the fraction of class-$i$ points from which class $j$ is reachable; inset: standard confusion matrix on unperturbed inputs.}
  \label{fig:torcm_example}
\end{figure}

Figure~\ref{fig:torcm_example} illustrates the idea on a toy problem with four classes in a two-dimensional feature space. 
The left panel shows the decision regions of a nearest-centroid classifier. 
For a point $\mathbf{x}^{(o)}$ and tolerance radius $\tau$ (as an example, $L_2$ distance is used), we define a tolerance region $\mathcal{T}_\tau(\mathbf{x}^{(o)})$ (dashed circle). 
For each target class $j$, we search within $\mathcal{T}_\tau(\mathbf{x}^{(o)})$ for a \emph{feasible point} $\mathbf{x}^*$ such that $f(\mathbf{x}^*) = j$. 
When such a point exists, we mark it by a cross in the left panel and connect it to $\mathbf{x}^{(o)}$, indicating that class $j$ is reachable from $\mathbf{x}^{(o)}$ under the tolerance budget.

The right panel shows the TOR-CM at a fixed shared tolerance radius $\tau=0.30$, computed by repeating the same reachability check over a collection of points from each true class. 
Each entry reports the fraction of class-$i$ points for which class $j$ is reachable within $\mathcal{T}_\tau(\mathbf{x}^{(o)})$. 
For example, for true class $0$, class $1$ is reachable for about $74\%$ of points and class $2$ for about $14\%$, while class $3$ is not reachable at this budget (entry $0.00$). 
Similarly, for true class $1$, classes $0$ and $3$ are frequently reachable (approximately $66\%$ and $29\%$). 
The inset shows the standard confusion matrix on the unperturbed points, which is perfectly diagonal in this toy setup. 
TOR-CM therefore complements standard evaluation by summarizing prediction stability in terms of feasible class transitions under bounded perturbations.

\subsection{TOR-CM for MNIST via distilled surrogates}
\label{sec:mnist_torcm}

To illustrate TOR-CM in an image setting, we evaluate four classifiers on MNIST and compute a TOR-CM for each:
Logistic Regression (LogReg), Decision Tree (CART), Random Forest (RF), and a CNN.
TOR-CM requires repeated counterfactual searches toward specified target classes.
To enable a unified gradient-based search for the non-differentiable classical models, we train smooth MLP \emph{surrogates} for LogReg, CART, and RF via knowledge distillation: each surrogate minimizes the KL-divergence to the teacher model’s predicted class-probability vector.
The CNN is inherently differentiable and is used directly for computing counterfactuals.
All neural models (surrogates and CNN) are implemented in PyTorch.

\paragraph{Counterfactual search and tolerance region}
For each test image $\mathbf{x}^{(o)}$ and each target class $j$, we run XA-$L_0$ to obtain an optimized image $\mathbf{x}^*$. 
For the classical models (LogReg, CART, RF), although the counterfactual is searched by XA-$L_0$ using a differentiable surrogate model, the target reachability is always confirmed using the original teacher classifier. 
We declare class $j$ reachable from $\mathbf{x}^{(o)}$ only if the prediction by the teacher model satisfies $f(\mathbf{x}^*)=j$.

The tolerance region is defined using the same interpretable penalty that governs XA-$L_0$ modifications, excluding the classification hinge term. 
We measure the tolerance loss by
\begin{eqnarray}
\mathcal{L}(\mathbf{x}',\mathbf{x}^{(o)})
=
\mathcal{L}_{\mathrm{XA\text{-}L_0}}(\mathbf{x}',\mathbf{x}^{(o)})
+
\mathcal{L}_{\mathrm{prox}}(\mathbf{x}',\mathbf{x}^{(o)}),
\label{eq:L_tolerance}
\end{eqnarray}
where $\mathcal{L}_{\mathrm{XA\text{-}L_0}}$ is the structured XA-$L_0$ penalty and $\mathcal{L}_{\mathrm{prox}}$ is the proximity (magnitude) penalty used by XA-$L_0$.

\paragraph{From $\lambda$-sweeps to reachability under a fixed budget $\tau$}
We adjust the objective function used by XA-$L_0$ in Eq.~\ref{L_comp} slightly: 
\begin{eqnarray}
&&\mathcal{L}_{\mathrm{comp}}(\mathbf{x}',\mathbf{x}^{(o)}) \nonumber \\
&=&\mathcal{L}_{\mathrm{cls}}(\mathbf{x}')
+
\lambda\big(
\mathcal{L}_{\mathrm{XA\text{-}L_0}}(\mathbf{x}',\mathbf{x}^{(o)})
+
\mathcal{L}_{\mathrm{prox}}(\mathbf{x}',\mathbf{x}^{(o)})
\big), \nonumber
\end{eqnarray}
where we use a single penalty weight $\lambda$ to be consistent with the tolerance loss in Eq.~\ref{eq:L_tolerance}. Since the tolerance loss is not directly set in optimization, we experiment with a set of $\lambda$ values taken from a predefined grid. 
For each pair $(\mathbf{x}^{(o)},j)$ and each $\lambda$, we obtain an output $\mathbf{x}^*(\lambda)$ and record the achieved tolerance loss $\mathcal{L}(\mathbf{x}^*(\lambda),\mathbf{x}^{(o)})$ together with a binary success indicator
$B(\lambda)=\mathbb{I}\{f(\mathbf{x}^*(\lambda))=j\}$.
We then define the cutoff value
\[
\mathcal{L}^*(\mathbf{x}^{(o)},j)
=
\min_{\lambda:\,B(\lambda)=1}
\mathcal{L}(\mathbf{x}^*(\lambda),\mathbf{x}^{(o)}),
\]
with the conventions that if $B(\lambda)=0$ for all $\lambda$ we set $\mathcal{L}^*$ to a value larger than all achieved losses, and if $B(\lambda)=1$ for all $\lambda$ we set $\mathcal{L}^*=\varepsilon$, where $\varepsilon$ is below all the values of $\tau$ examined in the experiments.
For any fixed tolerance budget $\tau$, we declare class $j$ reachable from $\mathbf{x}^{(o)}$ iff $\tau \ge \mathcal{L}^*(\mathbf{x}^{(o)},j)$.
Aggregating these reachability indicators yields TOR-CM $C_{i,j}$ as in~\eqref{eq:tor-cm}. 
For visualization, we report the per-class normalized matrix $\tilde C_{i,j}=C_{i,j}/|\mathcal{D}_i|$.

\paragraph{Choice of tolerance budgets}
To assess robustness across classifiers over a meaningful range of interpretable modification budgets, we report TOR-CM at three fixed tolerances, $\tau\in\{120,350,700\}$. 
These values are chosen to span three qualitatively distinct regimes of the counterfactual search: a \emph{low-budget} regime where only the easiest class transitions are feasible and TOR-CM remains close to diagonal; an \emph{intermediate} regime where off-diagonal reachability begins to appear for a subset of class pairs, exposing early differences in robustness across models; and a \emph{high-budget} regime where many more transitions become feasible, revealing how quickly each classifier’s reachable set expands under interpretable perturbations. 
Using a shared set of $\tau$ values across all methods enables direct comparison of robustness at the same absolute tolerance, while still capturing how stability degrades as the tolerance is relaxed.

\begin{figure}[H]
  \centering
  \includegraphics[width=\linewidth]{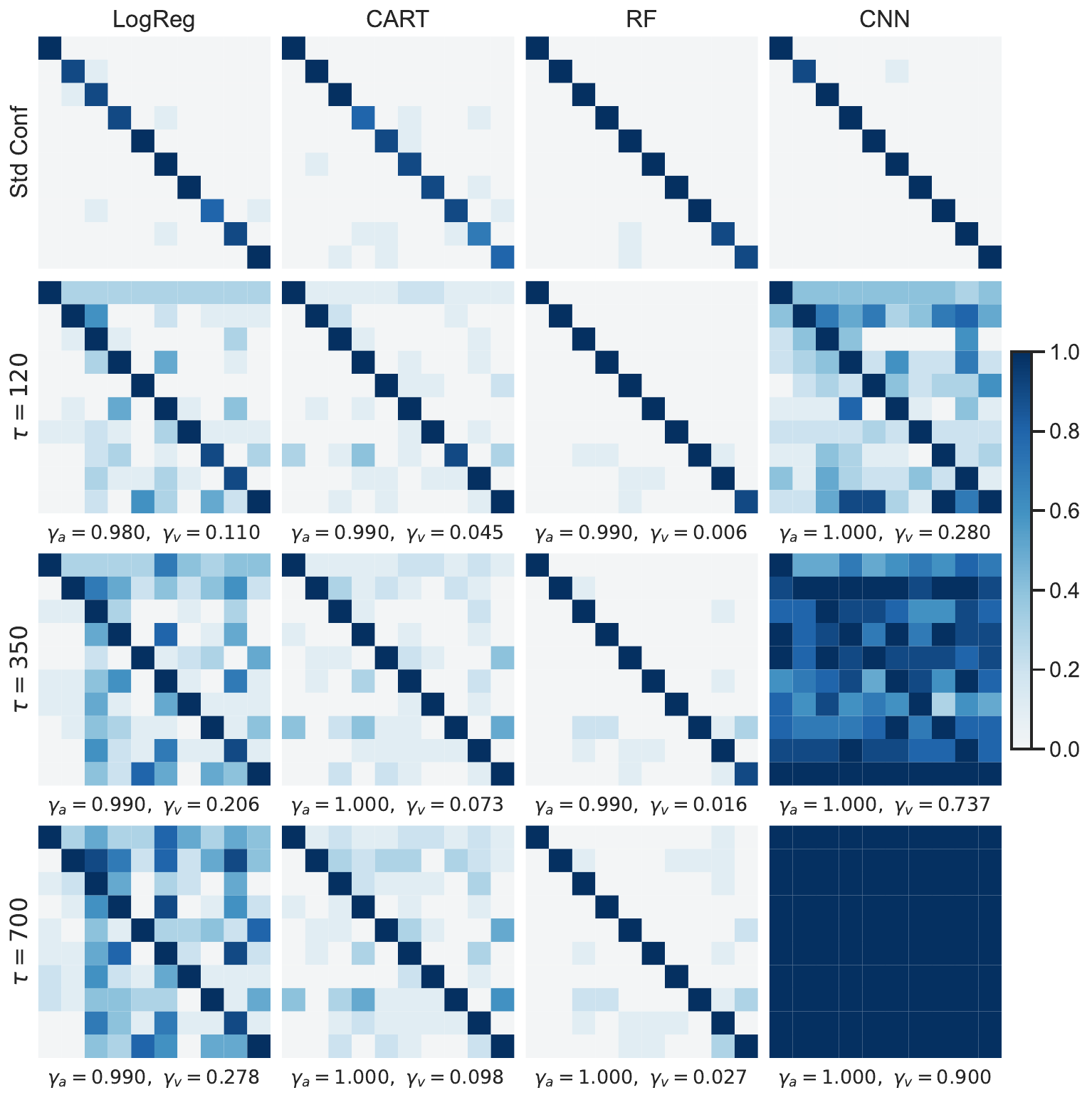}
  \caption{MNIST TOR-CM (per-class normalized) based on distilled surrogate models. The four columns of the plots correspond to the classification methods: LogReg, CART, RF, CNN. Top row: standard confusion matrices. Rows 2--4: TOR-CM at $\tau=120,350,700$ respectively. Beneath each panel, the values of  $\gamma_{\mathrm{a}}$ and $\gamma_{\mathrm{v}}$ are presented.}
  \label{fig:mnist_torcm_full}
\end{figure}

\paragraph{Results}
We evaluate on a balanced subset of $100$ MNIST test images (10 per class).
For each model and each tolerance budget $\tau\in\{120,350,700\}$, we compute TOR-CM and summarize robustness using $\gamma_{\mathrm{a}}$ and $\gamma_{\mathrm{v}}$ defined by~\eqref{eq:robust-metrics-accuracy}--\eqref{eq:robust-metrics-vul}.
Figure~\ref{fig:mnist_torcm_full} shows that TOR-CM reveals robustness differences that are largely invisible in the standard confusion matrices (top row). 
An interesting observation from this figure is that although the CNN achieves the highest accuracy on original instances, it is clearly less robust than the other methods. While achieving competitive accuracy on the original instances, RF and CART are the most robust across all tolerance levels, with RF outperforming CART.

At the smallest budget $\tau=110$, LogReg already exhibits noticeable off-diagonal reachability $\gamma_v=0.106$, whereas CART remains comparatively stable $\gamma_v=0.045$ and RF is the most stable $\gamma_v=0.006$. 
In contrast, the CNN shows substantially higher reachability even at this low budget with $\gamma_v=0.280$, indicating that many target classes can be achieved within the tolerance region.
As the budget increases to $\tau=350$, the difference between the models becomes much more pronounced. 
LogReg’s off-diagonal mass grows further $\gamma_v=0.206$, and CART increases modestly $\gamma_v=0.073$, while RF remains close to diagonal $\gamma_v=0.016$. 
The CNN, however, undergoes a sharp expansion in reachability, with a large fraction of class transitions becoming feasible, yielding $\gamma_v=0.737$, as shown by a substantially denser TOR-CM than all other methods.
At the largest budget $\tau=700$, LogReg continues to broaden its reachability ($\gamma_v=0.278$) and CART increases $\gamma_v$ slightly to $0.098$, while RF remains the most stable overall with $\gamma_v=0.027$. 
The CNN nearly saturates TOR-CM: any class becomes almost always reachable from any data point with the vulnerability indicator rising to $\gamma_v=0.900$. 
Across all methods, $\gamma_a$ is near 1 at moderate and large budgets, indicating that the true class is typically reachable within the tolerance region; the key differences are therefore driven by how rapidly the off-diagonal reachability grows as $\tau$ increases.
\section{Conclusions \& Discussion}\label{sec:conclusion}
We have presented a unified framework for generating interpretable counterfactuals and assessing the robustness of black-box classifiers. 
Central to our approach is the explainability-aware $L_0$ (XA-$L_0$) penalty, which encourages sparse and structurally coherent modifications, combined with a misclassification-driven loss that steers perturbed instances toward specified target labels. 
Our experiments on both tabular and image datasets demonstrate that XA-$L_0$ consistently produces compact, interpretable corrections while avoiding the diffuseness typical of proximity-only or unstructured-sparsity baselines. Furthermore, the Tolerance-Region Confusion Matrix (TOR-CM) provides a complementary tool for assessing robustness under interpretable modifications, revealing vulnerabilities that standard evaluation may miss. 

A natural direction for future work is to extend the method to datasets with categorical or mixed-type features. 
In this setting, performing counterfactual optimization is challenging, and determining which feature changes are interpretable is highly data-dependent.
Another promising avenue is to tailor the explainability-aware sparsity penalty to specific application domains, allowing it to encode practically motivated constraints or preferences. 
Furthermore, counterfactuals produced by the framework could inform improvements to the underlying classifier or guide targeted data collection, by revealing systematic patterns in misclassification. 
More broadly, it would be interesting to integrate this approach with active learning or model debugging pipelines, where interpretable counterfactuals not only explain errors but also suggest actionable interventions to enhance model robustness and fairness.
\bibliographystyle{IEEEtran}
\bibliography{ref}

\begin{IEEEbiography}[{\includegraphics[width=0.9in,height=1.25in,clip,keepaspectratio,trim=0 8 0 1]{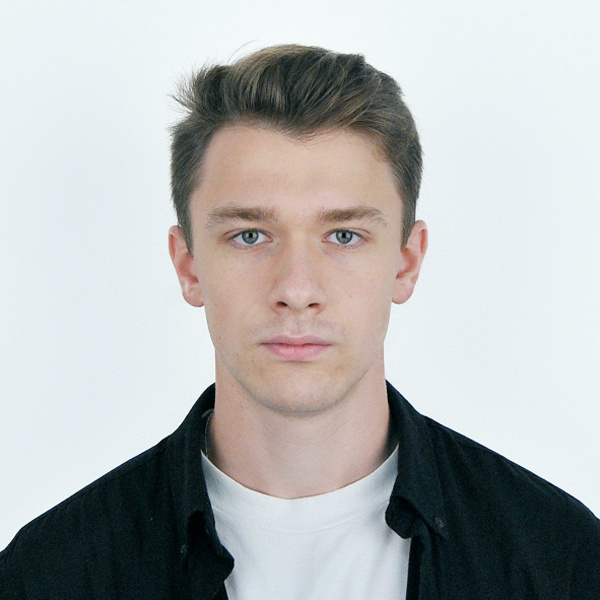}}] 
{Evgenii Kuriabov} is a Ph.D. candidate in the Department of Statistics at The Pennsylvania State University. His research interests include interpretable machine learning, counterfactual explanations, and optimization-based methods for diagnosing model errors and assessing robustness.

\end{IEEEbiography}

\vskip -20pt plus -1fil

\begin{IEEEbiography}[{\includegraphics[width=0.9in,height=1.25in,clip,keepaspectratio,trim=0 8 0 1]{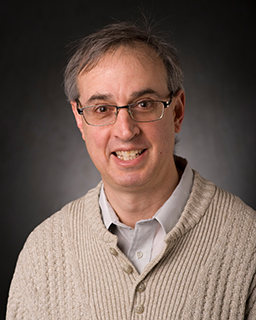}}] 
{David Miller} David J. Miller received his Bachelor’s from Princeton University in 1987, his M.S. from University of Pennsylvania, and his Ph.D. from UC Santa Barbara, all in electrical engineering. Dr. Miller joined Penn State’s EE Department in 1995. He is a longtime researcher in machine learning, data compression, and statistical estimation. He is an author of the 2023 Cambridge University Press book “Adversarial Learning and Secure AI”. He received an NSF CAREER Award in 1996. He was on the IEEE SP Society Conference Board from 2019-2022 and is currently on the Management Board for IEEE Transactions on Artificial Intelligence. He was Chair of the Machine Learning for Signal Processing Technical Committee, within the IEEE Signal Processing Society from 2007-2009. He was an Associate Editor for IEEE Transactions on Signal Processing from 2004-2007. He was General Chair for the 2001 IEEE Workshop on Neural Networks for Signal Processing.  
\end{IEEEbiography}

\vskip -20pt plus -1fil

\begin{IEEEbiography}[{\includegraphics[width=0.9in,height=1.25in,clip,keepaspectratio,trim=0 8 0 1]{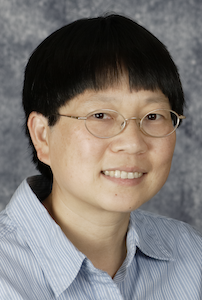}}] 
{Jia Li} (Fellow, IEEE) is a Professor of Statistics and (by courtesy) Computer Science and Engineering at The Pennsylvania State University. Her research interests include machine learning, artificial intelligence, probabilistic graph models, and image analysis. She worked as a Program Director at the NSF from 2011 to 2013, a Visiting Scientist at Google Labs in Pittsburgh from 2007 to 2008, a researcher at the Xerox Palo Alto Research Center from 1999 to 2000, and a Research Associate in the Computer Science Department at Stanford University in 1999. She received the MS degree in Electrical Engineering (1995), the MS degree in Statistics (1998), and the PhD degree in Electrical Engineering (1999), from Stanford University. She was Editor-in-Chief for Statistical Analysis and Data Mining: The ASA Data Science Journal from 2018 to 2020. She is a Fellow of the Institute of Electrical and Electronics Engineers and a Fellow of the American Statistical Association.
\end{IEEEbiography}

\end{document}